# Scaling Clinical Judgment to Evaluate Medical AI

Thomas A. Buckley[1], Zahir Kanjee[2], Peter G. Brodeur[2], Byron Crowe[3], Anthony M. Pettinato[2], Aashna P. Shah[1], Adrian D. Haimovich[4], Liam G. McCoy[2,5,6], Daniel Restrepo[7], Jason A. Freed[2], Ethan Goh[8,9], Jonathan H. Chen[3,8,9], Laura Zwaan[10], Katherine E. Goodman[11,12], Daniel J. Morgan[11,13], Raja-Elie E. Abdulnour[14], Adam Rodman[1,2]*, Arjun K. Manrai[1]*

1. Department of Biomedical Informatics, Harvard Medical School, Boston, MA
2. Department of Medicine, Beth Israel Deaconess Medical Center, Boston, MA
3. Stanford Division of Hospital Medicine, Stanford University, Stanford, CA
4. Department of Emergency Medicine, Beth Israel Deaconess Medical Center, Boston, MA
5. Division of Neurology, University of Alberta, Edmonton, Canada
6. Institute for Medical Engineering and Science, MIT, Cambridge, MA
7. Department of Medicine, Massachusetts General Hospital, Boston, MA
8. Stanford Division of Computational Medicine, Stanford University, Stanford, CA
9. Stanford Clinical Excellence Research Center, Stanford University, Stanford, CA
10. Institute of Medical Education Research, Erasmus Medical Center, Rotterdam, The Netherlands
11. Department of Epidemiology and Public Health, University of Maryland School of Medicine, Baltimore, MD
12. University of Maryland Institute for Health Computing, Bethesda, MD
13. VA Maryland Healthcare System, Baltimore, MD
14. Department of Pulmonary and Critical Care Medicine, Brigham and Women's Hospital, Boston, MA

*Co-senior authors
Correspondence: Arjun_Manrai@hms.harvard.edu

## ABSTRACT

Blinded physician evaluation has been considered by many to be the gold standard for assessing clinical reasoning in large language models (LLMs). This is difficult to scale; thus, prior studies typically rely on small physician panels, often from a single institution or specialty, which both limits the scientific questions investigated and makes it unclear whether findings would be reproduced with a different set of evaluators. To more rigorously and scalably study clinical reasoning in AI models, here we introduce PrecepTron, an LLM fine-tuned for physician-level evaluation of open-ended responses. PrecepTron was trained using low-rank adaptation (LoRA) of a 32-billion-parameter model on a small number of physician examples. We also release GRAND-ROUNDS, a new large-scale physician-annotated benchmark of 9,217 scored responses from 160 clinicians across seven studies. We show that frontier LLMs in typical “LLM-as-a-judge” approaches often disagree with physicians and with each other, but fine-tuning PrecepTron on a small number of cases enables physician-level consistent scoring across tasks. We use PrecepTron to reproduce headline findings from five influential studies assessing LLMs for clinical care in *JAMA*, *Science*, and *Nature Medicine* without new human grading. Using PrecepTron, we then pose new questions about how LLMs reason in medicine that would have been infeasible with human grading alone, including measuring the diagnostic accuracy of frontier LLMs when clinical cases are provided piecemeal, even token by token. Together, PrecepTron and GRAND-ROUNDS provide a foundation for reproducible, large-scale study of how LLMs reason in medicine. All code, data, and labels are made freely available for researchers.

## MAIN

Large language models (LLMs) have demonstrated remarkable capabilities across clinical reasoning tasks, offering opportunities to improve diagnosis and disease management[1–7]. As the models have gained capabilities, benchmarks have evolved from question-answer pairs to physician adjudication of largely open-ended LLM outputs[3,6–10]. While this approach maintains face validity[11] as models are being measured against a real-world comparator, it also has considerable challenges. First, it is slow, expensive, and resource-intensive given the need for expert human physician judges assessing potentially lengthy LLM outputs. This approach also has substantial methodological limitations and unresolved questions[12]. For instance, which experts are selected to serve as a human baseline: average physicians or expert physicians, generalists or subspecialists? How might geography, years in practice, community versus academic panels change the outputs? How reliable are the ratings from these experts? As benchmarking becomes essential for both health system deployment and national regulatory policy[13], these issues of construct and discriminant validity have the potential to become a reproducibility crisis, and could threaten both patient care and physician trust in AI.

Computational methods have been developed to attempt to increase the reliability and validity of LLM evaluations. Many studies now use LLMs themselves as automated graders of LLM outputs, a paradigm known as “LLM-as-a-judge”[10,14–18]. Such LLM-as-a-judge approaches are increasingly used to answer high-stakes questions, including whether medical specialization of LLMs that are widely used by practicing clinicians is necessary, or whether general purpose models outperform specialized tools[15]. Two problems limit this approach. First, even capable models can be poorly calibrated to human scores and carry systematic biases that might be worse than pooled physician outputs[19–25]. Moreover, it is unclear if LLM judges can reach physician-level agreement from a limited number of scored outputs. Second, using an externally hosted proprietary frontier model as a judge, as opposed to a local human physician, may be unacceptable to patients and health systems. As LLMs increasingly become deployed clinically, health systems need solutions for scalable oversight, but sending protected patient or institutional data outside the hospital system may be too great a risk.

To address these challenges with existing LLM-as-a-judge approaches, here we introduce PrecepTron, a fine-tuned LLM built on Qwen3-32B[26], trained using physician scores from multiple influential studies (Fig. 1). PrecepTron is capable of conducting scalable, expert-grade evaluation of clinical reasoning, calibrated to a specific physician panel from only a handful of scored examples per task. At 32 billion parameters, PrecepTron is small enough to run locally on hospital infrastructure, requiring only a single 80GB GPU for inference. To develop and validate PrecepTron, we assembled and openly release GRAND-ROUNDS (Graded Responses and Annotated Notes for Diagnostic Reasoning on UNstructured Data Sets), a new dataset of 9,217 physician scores spanning six clinical tasks, contributed by 11 grading physicians across seven published studies, over responses written by 160 clinicians and nine AI models. Benchmarking proprietary and open-source LLMs as automated evaluators on GRAND-ROUNDS, we find that current frontier models are often poorly aligned with physician scores when given only the published rubric and disagree even with one another. Fine-tuning PrecepTron on as few as two to a few dozen physician-scored cases per task closes this gap, reaching physician-level inter-rater agreement and matching or surpassing far larger frontier models.

To assess the validity and reliability of PrecepTron, we first use the system to reproduce five high-profile studies published in *JAMA*, *Nature Medicine*, and *Science* without any new human grading. We then demonstrate a use case of PrecepTron to unlock new scientific questions that would have been prohibitive if relying on high-cost physician grading alone. PrecepTron enables us to perform a large-scale experiment measuring how diagnostic and management accuracy change as an LLM receives clinical information on a token-by-token basis, across hundreds of cases and thousands of responses. This yields the surprising finding that frontier LLMs can solve challenging diagnostic and management cases using just the first few sentences of case text.

We openly release the GRAND-ROUNDS dataset and several variants of the trained PrecepTron-32B models on Hugging Face and at preceptron.net. We further release the protocols used to train PrecepTron so that other researchers can further refine the model using their data, if desired. PrecepTron may help reduce the cost of expert-grade reliable clinical evaluation, enhance the reproducibility of published clinical AI benchmarks, and unlock new scientific questions about how safe and trustworthy LLMs are in practice.

## RESULTS

### Frontier LLMs disagree with physicians and with each other

We evaluated eight LLMs — five open-source (Qwen3.5-9B, Qwen3-32B, Gemma-3-12B, Llama-3.1-8B, Mistral-3-8B) and three proprietary (Claude-Opus-4.6, Gemini-3.1-Pro, GPT-5) — as automated evaluators on the GRAND-ROUNDS benchmark across all five clinical tasks (Table 1, Fig. 2). We benchmarked each judge's agreement with physician scores against the agreement observed between physicians themselves (physician inter-rater agreement). The best proprietary model on each task reached physician inter-rater agreement on the *NEJM* Healer (Claude, 82% vs. 77% for physicians), BIDMC ER (Claude, 91% vs. 88%), and Landmark Diagnostic Cases (Gemini, 68% vs. 67%), but fell short on the *NEJM* CPCs (GPT-5, 87% vs. 92%) and the Grey Matters Management Cases (Claude, 84% vs. 95%). Agreement on the Landmark Diagnostic Cases varied widely across judges (28–30% for Claude and GPT-5 vs. 68% for Gemini). No single base model matched physician agreement across all five tasks. Agreement measured by quadratic-weighted κ, which corrects for chance agreement, showed the same pattern (Supplementary Fig. 1, Table 1), with a smaller gap between frontier models and physicians for the Grey Matters Management Cases. The open-source models were generally further from physicians on both metrics (Table 1).

The three proprietary judges disagreed not only with physicians but with one another, even when scoring the same responses against the same rubric (Fig. 2). Their mean assigned scores differed by up to 17 percentage points on the Landmark Diagnostic Cases (Gemini 74%, GPT-5 59%, Claude 57%) and by 9 points on the *NEJM* CPCs (GPT-5 80%, Gemini 77%, Claude 71%). These disagreements were not consistent within any one model. Claude was the harshest grader on the *NEJM* CPCs (mean score 71% vs. 88% for physicians) yet the most lenient on the BIDMC ER cases (82% vs. 78%). GPT-5 was the most generous of the three on the *NEJM* CPCs but scored the Landmark Diagnostic Cases 15 points below physicians (59% vs. 74%), while Gemini matched the physician mean on the Landmark cases (74%) but fell below physicians on the *NEJM* CPCs (77% vs. 88%) and above them on NEJM Healer (91% vs. 86%). Overall, a given model is calibrated to physician preferences on some tasks but not others, and which tasks differs by model. We also tested whether the three frontier judges favor their own model's outputs on four of the five tasks (all but BIDMC ER, whose patient case text cannot be sent to external APIs), and found no systematic self-preference (Supplementary Fig. 5).

**Lightweight LoRA fine-tuning aligns PrecepTron with physician scores**

We next asked whether low-rank adaptation (LoRA)[27] fine-tuning on a small number of physician-scored examples could close the gap between base-model judges and physicians. Trained with LoRA on just 2–46 cases per task, PrecepTron-32B improved over its Qwen3-32B base model on all five tasks, by 4 to 15 percentage points (e.g., 82% to 92% on the *NEJM* CPCs and 46% to 61% on the Landmark Diagnostic Cases), substantially closing the gap to physician inter-rater agreement (Fig. 3). PrecepTron-32B also performed on par with or outperformed the much larger frontier models across tasks. Its advantage was largest on the most demanding rubric: on the 19-point Landmark Diagnostic Cases, PrecepTron agreed with physicians on 61% of responses, outperforming GPT-5 (30%) and Claude (28%), but underperforming compared to Gemini (68%). We also compared PrecepTron to an ensemble of the three frontier models (GPT-5 + Claude + Gemini), finding that PrecepTron outperforms the ensemble on two of five tasks (*NEJM* CPCs, 92% vs. 83%; Landmark Diagnostic Cases, 61% vs. 44%) and matches it on the remaining three (Fig. 3). PrecepTron approaches physician inter-rater agreement, matching or exceeding it on three of five tasks (*NEJM* CPCs, *NEJM* Healer, and BIDMC ER). Agreement measured by Cohen's κ showed similar trends, with PrecepTron matching or exceeding physician inter-rater κ on the *NEJM* CPCs and approaching it on the *NEJM* Healer and BIDMC ER (Supplementary Fig. 1). The exception was BIDMC ER, where κ declined modestly after fine-tuning (0.68 to 0.60) despite higher accuracy. We include a confusion matrix in Supplementary Fig. 4 between physician and PrecepTron scores, showing high alignment and no collapse towards the majority class.

We conducted two ablation studies to assess the robustness of this training recipe. We first removed balanced resampling from the training regime. On the *NEJM* Healer corpus, quadratic-weighted κ fell from 0.72 to 0.51 despite similar accuracy, reflecting collapse toward the most common scores (Supplementary Table 1). We also conducted an ablation study where we removed LLM-generated examples from the training data for the Landmark Diagnostic Cases to determine if the model was shortcutting by systematically assigning higher scores to responses with LLM style (Supplementary Fig. 3), finding no evidence of shortcutting.

**Prompting fails to match fine-tuning**

We also compared LoRA fine-tuning to two prompt-based calibration strategies for the open-source judge (Qwen3-32B), using the same physician-scored training examples: few-shot prompting and an optimized prompt (GEPA) (Supplementary Fig. 2). Neither prompt-based strategy consistently improved on the base model. Few-shot prompting with five physician-scored examples improved some tasks while leaving others unchanged or lower (e.g., *NEJM* Healer, 76% to 67%). GEPA was similarly inconsistent and degraded management reasoning (Grey Matters Management Cases, 72% to 63%).

**PrecepTron reproduces five published studies without new human grading**

We next assessed whether PrecepTron could reproduce the findings of five prior influential studies that previously required extensive human grading (Fig. 4). For each study, we applied PrecepTron to score all responses using the original study's rubric and compared the resulting score distributions to those reported from physician evaluation. Cases used for training were excluded from this analysis. Across all five studies, PrecepTron reproduced the key headline findings reported in the original physician-graded analyses (Fig. 4), including the high diagnostic accuracy of GPT-4 from Kanjee et al. 2023, the relative ordering of chatbot and attending physician R-IDEA scores from Cabral et al. 2024, the finding from Goh et al. 2024 and Goh et al. 2025 that GPT-4 performed comparably to or outperformed physicians with conventional resources on diagnostic and management reasoning, and the pattern from Brodeur et al. 2026 of increasing diagnostic accuracy from triage to admission in the BIDMC emergency department study, with o1 performing on par with internal medicine physicians at each touchpoint.

**PrecepTron reveals that LLMs solve many cases from a small fraction of the case text**

Using PrecepTron as an automated grader, we conducted a large-scale experiment to measure how LLM performance evolves when provided a clinical case token-by-token. For each case, we provided GPT-5 with incrementally longer aliquots of case text and evaluated each response using the original rubric. We applied this approach across three tasks: *NEJM* CPCs, Grey Matters Management Cases, and Landmark Diagnostic Cases. For Grey Matters, we used only the five held-out questions that do not depend on case text revealed in later questions. To guard against memorization, the *NEJM* CPCs used here were 50 cases published from 2025 onward, after the pretraining cutoff of both GPT-5[28] and Gemma-4[29]. The Landmark Diagnostic Cases and Grey Matters Management Cases have not been released on the internet at the time these experiments were conducted.

Across all tasks, models could effectively solve cases with a very small amount of text (Fig. 5A), especially under the lenient grading metrics used in many publications (top-10 DDx accuracy). For example, on the *NEJM* CPCs, GPT-5 solved 42% of cases with 50 or fewer tokens of case text (approximately 40 words) when crediting a top-10 differential, falling to 14% for a top-3 differential and 12% for a top-1 diagnosis. For reference, the typical Presentation of Case of an *NEJM* CPC contains approximately 1,600 tokens (approximately 1,080 words). Management performance saturated even earlier: GPT-5 reached at least 80% of the rubric maximum on four of the five questions across the three held-out Grey Matters cases within the first 50 tokens, three of them within the first 10 tokens. Diagnostic reasoning behaved similarly, with all four Landmark Diagnostic Cases reaching at least 80% of the rubric maximum by 50 tokens.

We include representative cases in which GPT-5 reached a high score from 50 or fewer tokens of case text in Table 2. Three of the four CPC cases shown, each assigned a Bond score of 5 by PrecepTron-32B, contain the exact diagnosis in GPT-5's top-3 differential from 10 to 30 tokens of case text. The fourth ("A 70-year-old woman...") is a borderline judgment by PrecepTron-32B, potentially reflecting the leniency of the physician grading the judge was trained to reproduce. For the Landmark cases, high scores reflect the rubric's credit for reasoning quality rather than the final diagnosis alone.

This pattern generalized to a smaller open-source model: Gemma-4-31B showed the same qualitative behavior (Fig. 5B), solving many cases from short prefixes though requiring somewhat more text than GPT-5 to reach the same solve rate. We hypothesize that the especially rapid saturation observed in management tasks could reflect the structure of these questions, where key clinical details are frequently embedded directly in the prompt. Additionally, the rubrics used in prior studies may reward good guesses rather than true clinical reasoning, allowing models to game the rubric.

## DISCUSSION

We introduce PrecepTron, a fine-tuned LLM judge for automated, expert-grade evaluation of open-ended clinical reasoning, and GRAND-ROUNDS, a large-scale physician-annotated benchmark of clinical LLM outputs consisting of 9,217 physician scores across six clinical tasks. We find that frontier LLMs prompted only with published rubrics disagreed with one another and were poorly aligned with expert physician evaluators. By contrast, after lightweight fine-tuning of a weaker open-source model on a small number of physician-scored examples per task, PrecepTron reached physician inter-rater agreement and matched or surpassed substantially larger frontier models. At 32 billion parameters, PrecepTron is small enough to run locally within an academic medical center, enabling research without having to send patient or institutional data to external organizations. On real emergency-department cases from Beth Israel Deaconess Medical Center, PrecepTron achieved concordance with physician evaluators on par with the physicians themselves. PrecepTron also reproduced the headline findings of five previously published physician-graded studies. Finally, by enabling an evaluation experiment that would be impractical with human graders, PrecepTron revealed that GPT-5's scored diagnostic and management performance saturates within the first one to two sentences of case text.

Prior studies have trained LLM-as-a-judge for narrow clinical AI tasks, such as scoring OSCEs[30], clinical hallucination detection[31], or the quality of clinical summaries[24]. However, these prior works either had an ample number of labeled examples per task, or relied on synthetically generated data to train a judge model. For many real-world clinical tasks, labeled data is scarce due to the laboriousness of physician scoring, and data can be heavily imbalanced towards higher scores (in the case of a well-performing AI system). We introduce in this paper an effective method of calibrating a small LLM judge from limited labels: using 93 physician-scored responses spanning just 2 cases, we improve judge performance on the landmark diagnostic task from 46% to 61% agreement (κ from 0.62 to 0.80). We demonstrate that a simple balanced resampling of the training data enables high agreement with physician panels by preventing the model from collapsing to predicting the majority class, and we hypothesize that a LoRA fine-tune run for a small number of epochs limits how far the model weights can move, protecting against overfitting and catastrophic forgetting. This approach remains effective when potential shortcuts, such as AI-written responses, are removed from the training data. To make this method directly usable, we distill it into a stepwise protocol for deploying an LLM judge on a new clinical task (Box 1) and release the accompanying fine-tuning code, enabling researchers with a new task and limited physician data to train a calibrated judge model for scoring at scale.

Both very small physician panels and simple LLM-as-a-judge approaches are being increasingly used to answer high-stakes questions about the safety and trustworthiness of LLMs in clinical practice, with limited investigation into the reproducibility or reliability of such evaluations[12]. Our findings suggest that prior LLM studies consisting of small physician panels or simple LLM-as-a-judge approaches, even with published open rubrics specifying the formal scoring criteria, can be difficult to reproduce by a different set of graders. Our findings demonstrate that the latent preferences of a physician panel can be effectively learned, and used to reproduce the findings from five major clinical AI papers. PrecepTron therefore provides a way to study how new models would have performed against prior baselines, while remaining calibrated to the original physician panel and reducing variance introduced by new panels. This same technique could allow rater preferences to be "swapped out," effectively allowing researchers to re-appraise prior conclusions by, for example, swapping generalist physicians for specialists, or localizing a national benchmark by annotating a small set with local physicians. Furthermore, researchers can interrogate latent biases in prior physician evaluations by learning panel-specific preferences and studying them at scale, allowing for science not possible before because of the scale of human annotations necessary.

PrecepTron uncovered that LLMs can sometimes solve diagnostic and management dilemmas using small text aliquots of the case presentation, raising thorny questions about what current clinical LLM benchmarks actually measure. GPT-5 reached at least 80% of the rubric maximum on four of the five Grey Matters questions within 50 tokens of case text, most within 10 tokens. It did the same on all four Landmark Diagnostic Cases within 50 tokens and produced a top-10 differential containing the correct diagnosis for 42% of CPC cases from 50 or fewer tokens. These cases were not available in pretraining data, suggesting that the result is not explained by memorization. Additionally, this is not completely explained by a failure of PrecepTron being overly lenient to incorrect responses (see examples in Table 2). Instead, we hypothesize that models can often map a small number of demographic and symptom tokens onto a prototypical presentation and generate a differential diagnosis or workup that is sufficient to score well under existing rubrics. Pattern recognition is an important component of clinical reasoning. However, these findings suggest that current benchmarks may reward early pattern matching while leaving untested whether a model does the essential cognitive clinical tasks of integrating new information, revising an initial impression, or avoiding premature closure as a case unfolds. Future evaluations should reassess outcomes and vary the information available to the model over time and credit appropriate updating, rather than relying only on a single end-of-case score.

The datasets, models, and fine-tuning algorithm we release are intended to enable new research questions about clinical AI. First, researchers studying a new clinical task can follow the stepwise protocol we provide (Box 1) to calibrate a judge model to a small number of physician-scored examples, enabling larger-scale evaluation without requiring exhaustive human grading. Second, GRAND-ROUNDS provides a large, heterogeneous collection of physician-scored model outputs that can be used to study systematic biases in LLM judges, including preferences for longer or more detailed responses, models from the same family, or AI- versus human-written responses. Third, the release includes physician annotations from multiple influential clinical AI studies that were not previously available as reusable benchmarks. Researchers can therefore evaluate new models against these prior baselines using judges calibrated to the scoring preferences of the original physician panels, rather than convening a new panel whose latent preferences may differ. Finally, because PrecepTron can be deployed locally, these calibrated judges can be used within academic medical centers to evaluate model outputs generated from protected clinical data. More broadly, the same approach can be adapted to other clinical evaluation tasks for which a rubric exists but only a small number of physician scores can feasibly be collected.

This study has several limitations. Physician agreement was modest for some GRAND-ROUNDS tasks. These values both upper-bound achievable judge–physician agreement and reflect genuine variation among clinicians on nuanced reasoning tasks. In addition, because PrecepTron is calibrated to each source study's physician panel, it inherits that panel's scoring preferences and biases. Reproducing a study with PrecepTron therefore reproduces that study's evaluation standard; it does not independently re-adjudicate the underlying clinical judgments, and consensus across panels remains a separate question to be answered via traditional social science methods. The judges were also trained on small datasets by design. This supports practical deployment but may leave rubric edge cases underrepresented. Finally, although PrecepTron recovered original score distributions and group-level orderings, it did not recreate the randomized comparisons or human–AI interaction protocols of the source studies. Physician evaluation will remain essential for such settings, particularly when measuring clinical impact, workflow integration, and human decision-making.

Overall, we find that existing LLM-as-a-judge approaches often fall short, disagreeing with physicians and with one another in ways that threaten the reproducibility of medical AI research. We show that lightweight fine-tuning on a small number of physician-scored examples using a relatively small model can substantially improve alignment with a target physician panel, enabling reproduction of prior high-impact findings and evaluation at a scale that would be impractical with human grading alone. Together, PrecepTron and GRAND-ROUNDS provide a practical framework for more rigorous, scalable, and reproducible evaluation of clinical AI.

As automated evaluation becomes more common, however, the central challenge will not simply be choosing a better judge, but establishing shared expectations for how judges are validated, reported, and interpreted. Future studies should make clear how closely LLM judges reproduce expert human assessments, where they fail, and how sensitive conclusions are to the choice of judge. Building consensus around these practices will be essential if results from medical AI evaluations are to be trusted by readers, compared across studies, and ultimately inform how clinicians should use clinical AI.

# METHODS

## Benchmark Construction (GRAND-ROUNDS)

The GRAND-ROUNDS benchmark was assembled from seven previously published physician-evaluated studies of open-ended clinical large language model (LLM) outputs. It includes 9,217 physician scores across 5,250 unique response–rubric entries; responses were written by 160 clinicians and 9 AI models and scored by 11 physician evaluators. The benchmark comprises six clinical tasks, five of which are used in all judge experiments:

1. ***NEJM* CPCs diagnosis:** differential-diagnosis quality on *NEJM* Clinicopathologic Conference (CPC) cases scored with the Bond score[4,9].
2. **Landmark Diagnostic Cases:** 19-point diagnostic reasoning rubric from Goh et al. 2024[2].
3. **Grey Matters Management Cases:** management reasoning on the Grey Matters vignettes from Goh et al. 2025[3] Each case presents a clinical scenario with multiple questions, and each question is scored against its own rubric.
4. ***NEJM* Healer:** consultation-note quality scored with the four-component R-IDEA rubric from Cabral et al. 2024[5].
5. **BIDMC ER:** emergency-department triage accuracy on the BIDMC ER dataset from Brodeur et al. 2026[6], scored with the Bond rubric[9]. The BIDMC ER cases and scores are not included in the public release for data privacy.

The sixth task, testing-plan scoring of the *NEJM* CPC cases (131 entries scored with the 0–2 CPC Testing Plan rubric), is included in the data release but excluded from the judge experiments because of its heavy skew towards high scores. Source studies comprised Kanjee et al. 2023[4], Cabral et al. 2024[5], Goh et al. 2024[2], Goh et al. 2025[3], Buckley et al. 2025[32], Brodeur et al. 2026[6], Buckley et al. 2026[33], and additional unpublished physician scores from our group. When multiple physicians scored the same entry, the entry-level physician score was taken as the reconciled or final score when one was recorded, and otherwise as the mean of the non-reconciled graders' scores. For the first time, we release case data for the Grey Matters management cases and the landmark diagnostic cases. The harmonized benchmark has been released as a public Hugging Face dataset (https://huggingface.co/datasets/tbuckley/GRAND-ROUNDS).

**Task Rubrics**

1. **Bond Score:** A five-point ordinal scale (0, 2, 3, 4, 5) used to assess diagnostic accuracy. A score of 5 indicates that the correct diagnosis is included in the differential[9].
2. **Landmark Diagnostic Cases Rubric:** A 19-point rubric that evaluates structured diagnostic reasoning. It allocates points for proposing candidate diagnoses with supporting and opposing evidence, identifying the correct final diagnosis, and suggesting appropriate next steps[2].
3. **Grey Matters Management Cases Rubric:** A case-specific, multi-criteria scoring system applied to management decisions. Scores for individual questions within a case are summed and normalized by the total possible points, yielding a continuous score between 0 and 1[3].
4. ***NEJM* Healer R-IDEA Rubric:** A 10-point rubric composed of four components: interpretive summary, differential diagnosis, explanation of the leading diagnosis, and explanation of alternative diagnoses. Each component is scored on a predefined subscale[5,34].
5. **CPC Testing Plan:** A three-point scale (0–2) used to evaluate proposed diagnostic workups. A score of 1 reflects a clinically reasonable plan that differs from the reference standard, while a score of 2 reflects a plan that matches or appropriately extends it[6].

**Base-model LLM-as-a-judge evaluation**

Eight LLMs were evaluated as zero-shot automated judges on the GRAND-ROUNDS test set: five open-source models (Qwen3.5-9B, Qwen3-32B, Gemma-3-12B, Llama-3.1-8B, Mistral-3-8B) and three proprietary models (Claude-Opus-4.6, Gemini-3.1-Pro, GPT-5). We additionally evaluated an ensemble of the three proprietary judges. For each scored entry, the ensemble score was the mean of the three judges' scores, rounded to the nearest valid score on the task's rubric scale. Locally served judges (Qwen3-32B and PrecepTron-32B via vLLM) were run using temperature 0, while all other judges were accessed via OpenRouter using default temperature. All judges used a maximum of 16,384 completion tokens and up to two retries. Judge outputs were required to be valid JSON containing a numeric score and a free-text justification; responses were parsed with a JSON-first, regex-fallback parser. Missing values were replaced with the value farthest from the physician score. Scores exceeding the rubric maximum were clipped to the rubric range. Prompts are available in the associated GitHub repository.

**Training PrecepTron-32B**

PrecepTron is trained by supervised fine-tuning (SFT) of Qwen3-32B. A separate adapter was fine-tuned for each rubric; the *NEJM* CPCs and BIDMC ER tasks share one adapter because they share the Bond rubric.

Train and test splits were constructed by case. For tasks with at least 10 cases, 20% of cases were used for training; for tasks with fewer than 10 cases (the Landmark Diagnostic Cases and the Grey Matters Management Cases), 2 cases were used for training. The Grey Matters Management Cases comprise five cases, each with several standardized questions; two cases were used for training and the remaining three cases were held out. The *NEJM* CPCs and BIDMC ER tasks, which share the Bond rubric, were pooled for training. The same 20% training split was used for the prompt-based baselines described below.

SFT was performed on a single NVIDIA H100 80 GB GPU using low-rank adapters (LoRA) applied to all query, key, value, output, gate, up-, and down-projection matrices in every transformer block, with rank 16, scaling factor 32, and dropout 0.05. Training used the AdamW optimizer under a cosine learning-rate schedule (peak learning rate $2 \times 10^{-4}$, 3% warm-up), a per-device batch size of 1 with 4 gradient-accumulation steps (effective batch size 4), a maximum sequence length of 4,096 tokens, and bfloat16 precision. The supervised target for each training example was the physician reconciled score. When there were multiple physician scores for an example but no reconciled score, a random score was selected. Models are trained with the same system and user prompt the judge receives at inference time. Loss was computed on the assistant tokens only.

To prevent skewed physician-score distributions from biasing the judge toward the majority score region, training entries were oversampled with bin-stratified resampling applied uniformly to every task. Each entry was binned by its normalized physician score (physician score divided by the rubric maximum) into terciles of the rubric scale — low [0, 1/3), mid [1/3, 2/3), and high [2/3, 1.0]. The largest bin's count was taken as the target, and each smaller bin's entries were repeated (with reshuffling so the same example does not appear in consecutive gradient steps) until the bin reached the target, capped at five times the bin's original size so that a handful of rare examples could not dominate the gradient.

We tested for shortcutting by retraining the Landmark Diagnostic Cases adapter on human-authored responses only, with all AI-model responses removed from training, and comparing its scores on held-out responses to those of the standard adapter (Supplementary Fig. 3). We also swept epoch counts to check for overfitting (Supplementary Table 1). For the released version of PrecepTron-32B, and the version shown in Table 1, we trained for 3 epochs with balanced resampling, using a fixed seed for reproducibility.

**Prompt-based baselines (few-shot and GEPA)**

We compared SFT against two baselines that align an LLM judge with physician preferences through the prompt alone, with no weight updates, using the same 20% by-case training split as SFT.

First, we used few-shot prompting[35], where we include five physician-scored examples drawn from the training split in the prompt. Second, we used a state-of-the-art prompt optimizer, GEPA[36], which discovers effective prompts by optimizing a metric over a training set. For GEPA, the 20% training split was further divided by case into a training subset and a held-out validation subset. The metric optimized by GEPA was the normalized score: $1 - |s_LLM - s_physician| / s_max$. Each GEPA run explored up to 15 candidate prompts and selected the candidate that maximized alignment on the held-out validation subset.

**Statistical Analysis**

The primary metric was accuracy against the reconciled ground-truth physician score (or the mean score when no reconciled score was available): within 1 point on the raw 0–5 Bond scale for the NEJM CPCs and BIDMC ER tasks, and within 10% of the normalized score for the remaining tasks. For the Grey Matters management task, question-level scores were summed to the case level before computing accuracy; all other tasks were scored per entry. The secondary metric was quadratic-weighted Cohen's κ, computed at the level of individual scored entries for every task after rounding each score to the nearest integer. Physician inter-rater agreement was computed on the subset of entries scored by at least two non-reconciled physicians. Ninety-five-percent confidence intervals for both metrics were generated by non-parametric bootstrap over the scored units (entries, or summed cases for management accuracy) with 2,000 resamples and a fixed random seed.

## CODE AND DATA AVAILABILITY

All code for training and evaluating PrecepTron, including the LoRA fine-tuning recipe, judge prompts, few-shot and GEPA baselines, and scripts to reproduce the analyses in this manuscript, is available at https://github.com/2v/PrecepTron. Trained PrecepTron-32B adapters are available on Hugging Face at https://huggingface.co/collections/tbuckley/preceptron. The GRAND-ROUNDS benchmark is available at https://huggingface.co/datasets/tbuckley/GRAND-ROUNDS. The release contains 7,395 physician scores across 4,339 entries. The Landmark Diagnostic Cases are included as responses, physician scores, and final diagnoses; their case text is not released. The BIDMC emergency-department dataset contains protected health information and is not included. Documentation and a project overview are available at https://preceptron.net.

learning. *arXiv [cs.CL]* (2025).

## Fig. 1: PrecepTron enables physician-level evaluation of AI outputs at scale

### A. PrecepTron for Physician-level Evaluation of AI Outputs

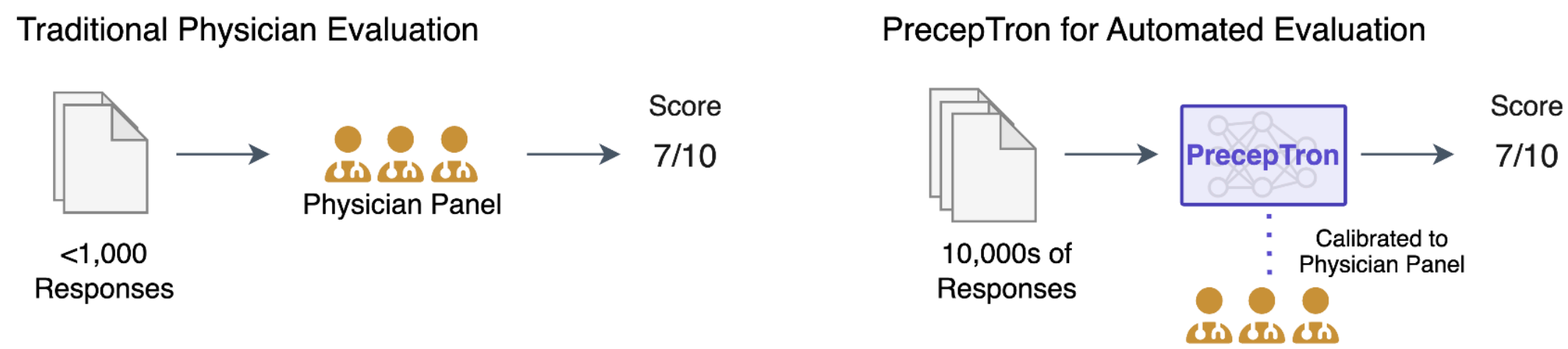

### B. Creating GRAND-ROUNDS, A Dataset for Training and Validating AI Evaluators

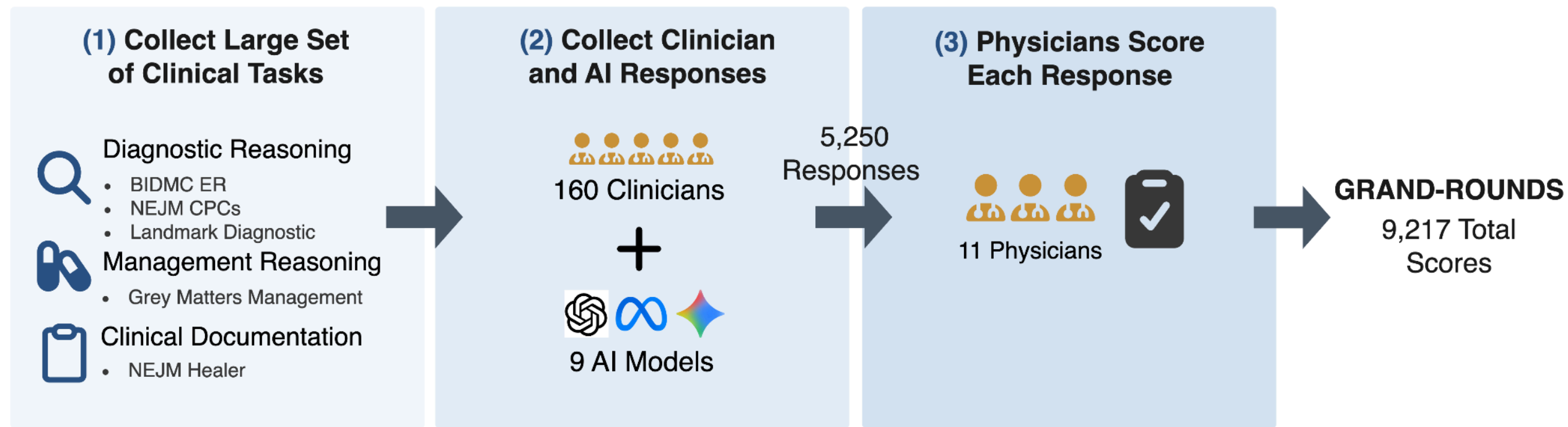

### C. Training PrecepTron to be Calibrated to a Small Amount of Physician Scores

### D. PrecepTron Reproduces the Core Findings of 5 Published Papers

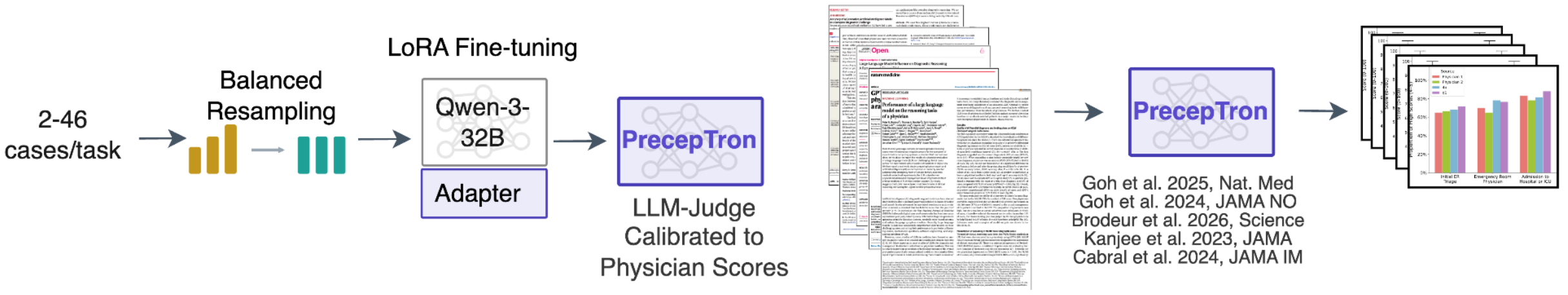

**Fig. 1:** PrecepTron enables physician-level evaluation of open-ended clinical AI outputs. **A.** Evaluating open-ended clinical LLM responses traditionally requires a panel of physicians to independently score each response against a validated rubric. This manual scoring is the central bottleneck to studying clinical AI models and to reproducing published findings, limiting the scale at which either can be done. PrecepTron replaces the physician panel with an LLM judge, calibrated to physician scores, that produces the same rubric-grounded scores automatically and at a scale manual grading cannot reach. **B.** GRAND-ROUNDS, the physician-annotated benchmark used to develop and validate PrecepTron, was built in three steps: collecting clinical tasks spanning diagnostic reasoning (BIDMC ER, NEJM CPCs, landmark diagnostic cases), management reasoning (Grey Matters), and clinical documentation (*NEJM* Healer); gathering 5,250 responses from 160 clinicians and 9 AI models; and having a panel of 11 physicians score each response, yielding 9,217 physician scores. **C.** A 20% training split (by case) is used to LoRA fine-tune Qwen3-32B, yielding a judge calibrated to physician scores. Each task-specific judge is trained on only 2 to 46 unique cases per task. **D.** Without any new human grading, PrecepTron reproduces the core findings of five published studies (Kanjee 2023; Cabral 2024; Goh 2024; Goh 2025; Brodeur 2026).

## Fig. 2: AI agreement with physicians and with each other

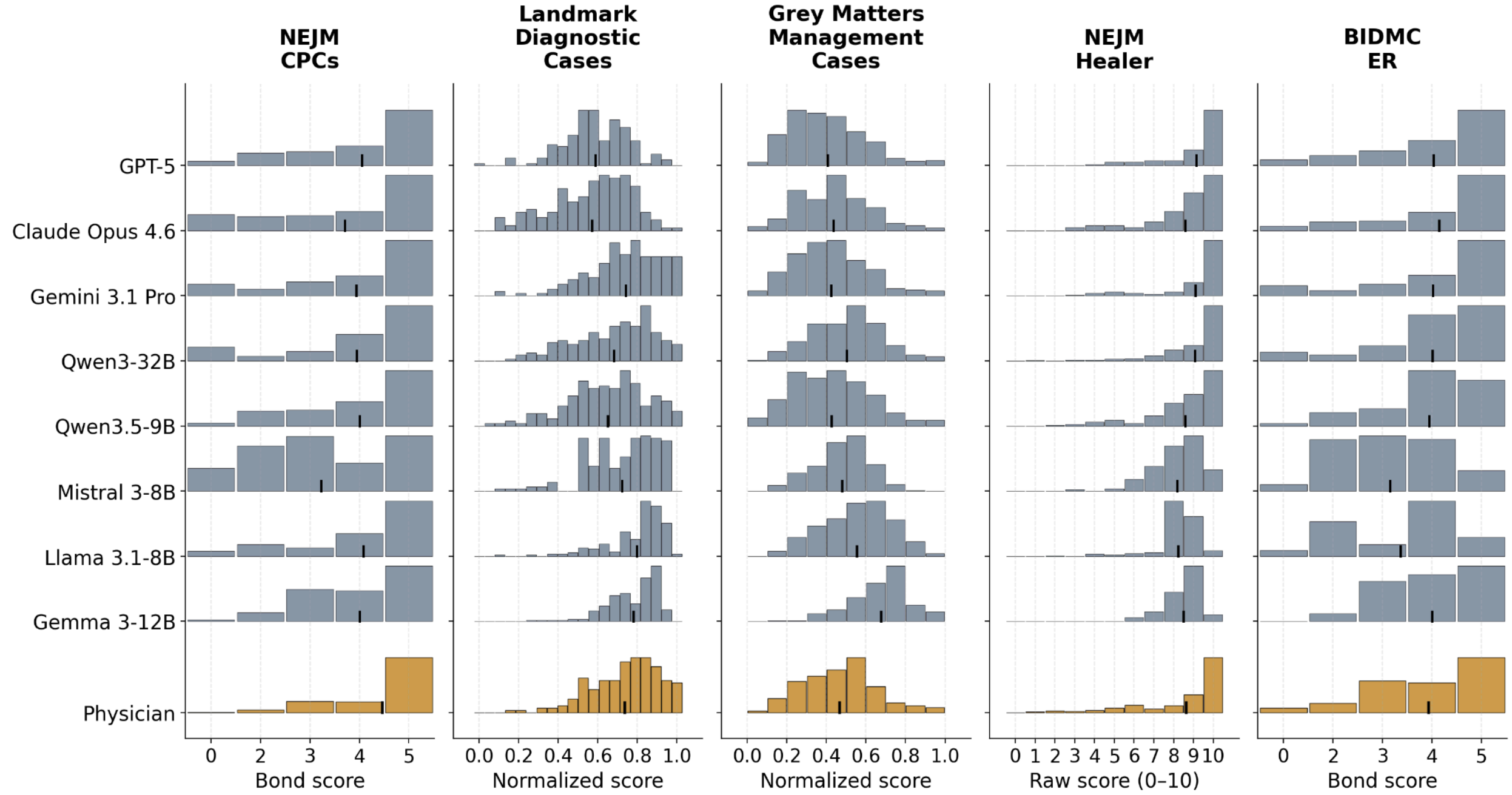

**Fig. 2:** Distribution of rubric scores assigned by AI judges versus physicians across five tasks. Each panel shows, for one GRAND-ROUNDS task, the distribution of scores assigned to the same set of responses by eight AI judges (blue) and by physicians (gold, bottom row). Tasks scored on the 0–5 Bond scale (NEJM CPCs, BIDMC ER) and the 0–10 R-IDEA scale (NEJM Healer) are shown as raw scores; landmark diagnostic and Grey Matters management scores are normalized to 0–1. Black ticks mark per-model means.

## Fig. 3: PrecepTron matches physician-level agreement across tasks

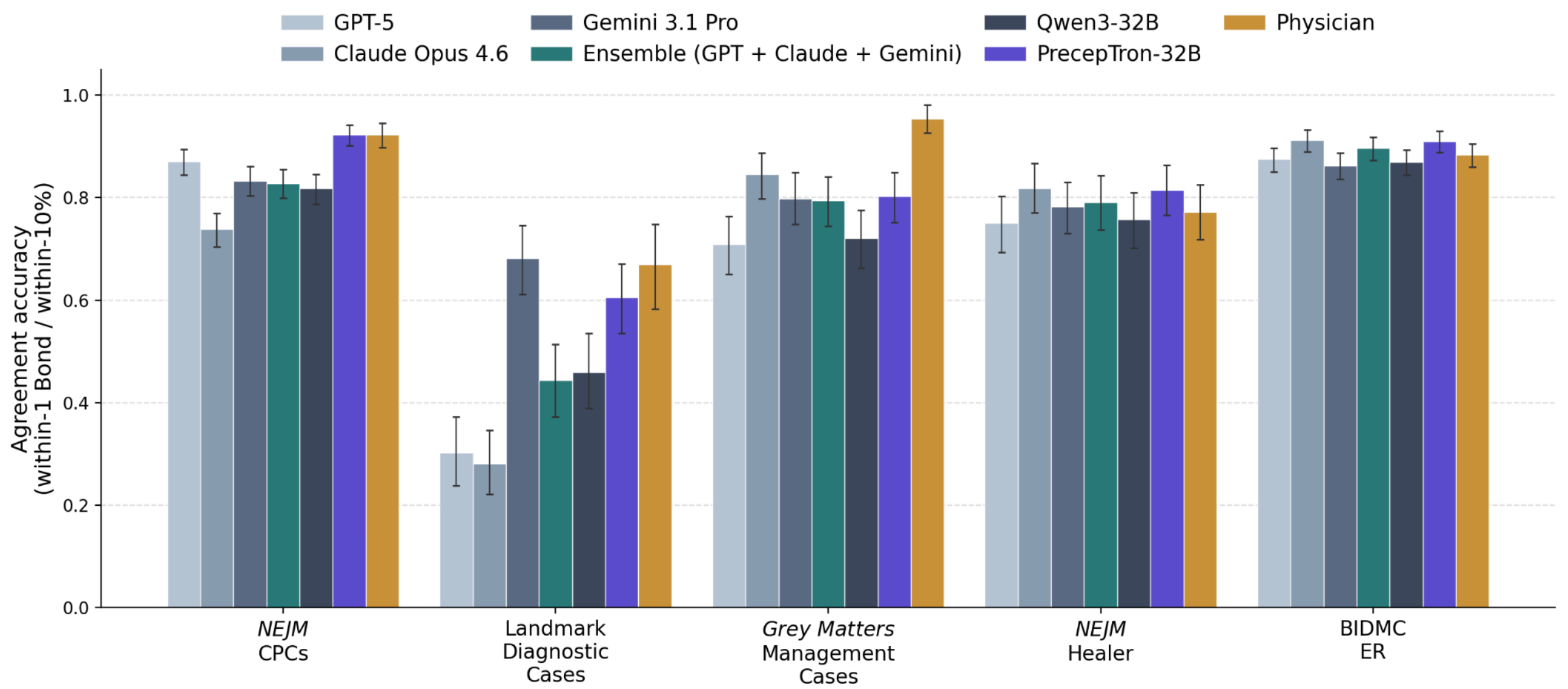

**Fig. 3:** Accuracy against physician scores for three proprietary models (GPT-5, Claude Opus 4.6, Gemini 3.1 Pro), an ensemble of those three models, the open base model (Qwen3-32B), the fine-tuned PrecepTron-32B, and the physician–physician baseline (gold), across the five GRAND-ROUNDS tasks. Agreement is the fraction of responses within 1 point on the 0–5 Bond scale for NEJM CPCs and BIDMC ER and within 10% of the normalized score elsewhere. The physician baseline is computed on the subset of cases scored by at least two physicians. Error bars are 95% bootstrap confidence intervals.

## Fig. 4: PrecepTron reproduces five published studies with no new human grading

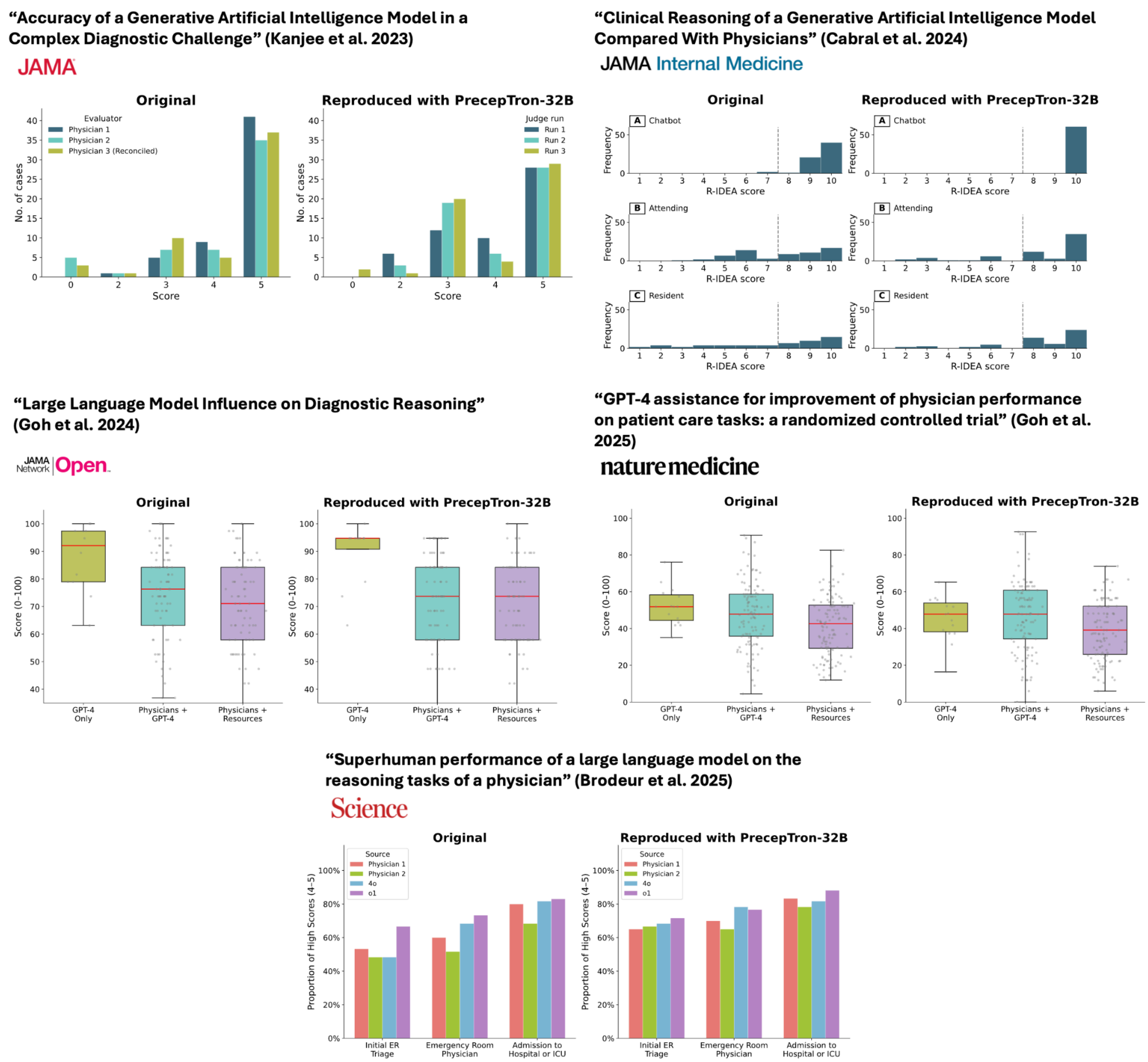

**Fig. 4:** PrecepTron reproduces the findings of five published studies without new human grading. For each study, every LLM and physician response from the original publication was re-scored with PrecepTron-32B using that study's original rubric and compared to the published physician-graded result (cases used for PrecepTron training were excluded). Top left: Bond-score histograms from Kanjee et al. (JAMA 2023), GPT-4 on NEJM CPC cases. Top right: R-IDEA score histograms from Cabral et al. (JAMA Internal Medicine 2024), comparing chatbot, attending, and resident consultation notes. Middle left: diagnostic-reasoning boxplots from Goh et al. (JAMA Network Open 2024), physicians with and without LLM assistance versus the LLM alone. Middle right: management-reasoning performance with and without GPT-4 assistance from Goh et al. (Nature Medicine 2025). Bottom: diagnostic accuracy across the BIDMC emergency-department triage (initial triage, ER evaluation, admission) from Brodeur et al. (Science 2026), comparing physicians and LLMs. In each panel the original physician-graded result (left) is shown alongside the PrecepTron-scored result (right).

## Fig. 5: GPT-5 and Gemma-4-31B solve many cases from a fraction of the case text

### A. GPT-5

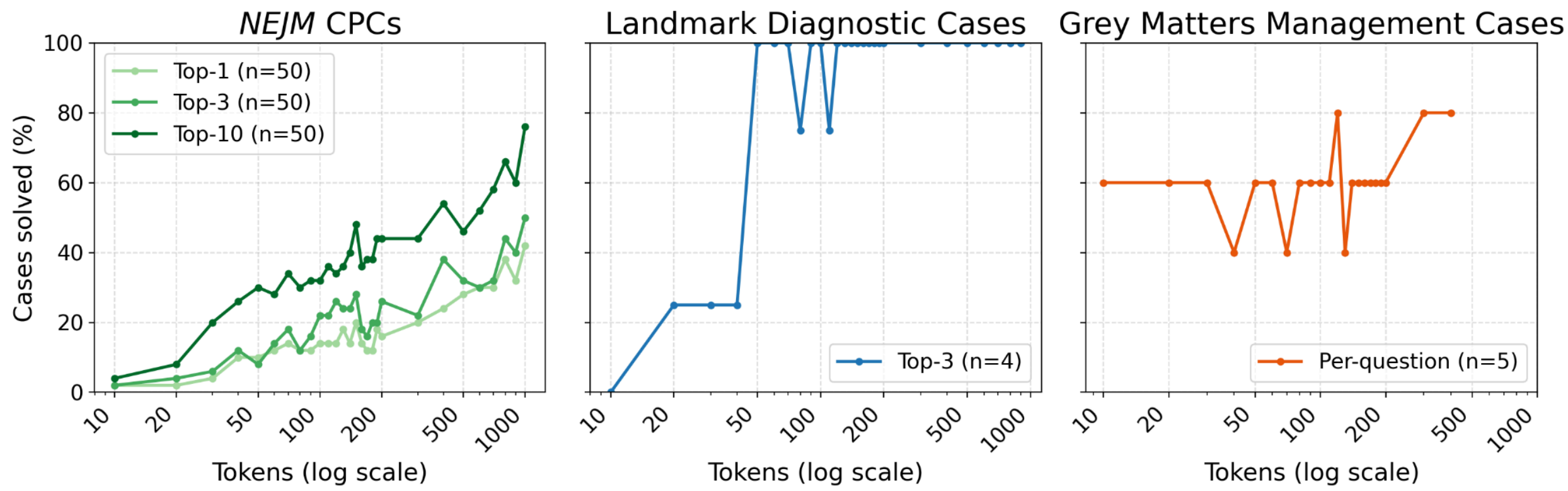

### B. Gemma-4-31B

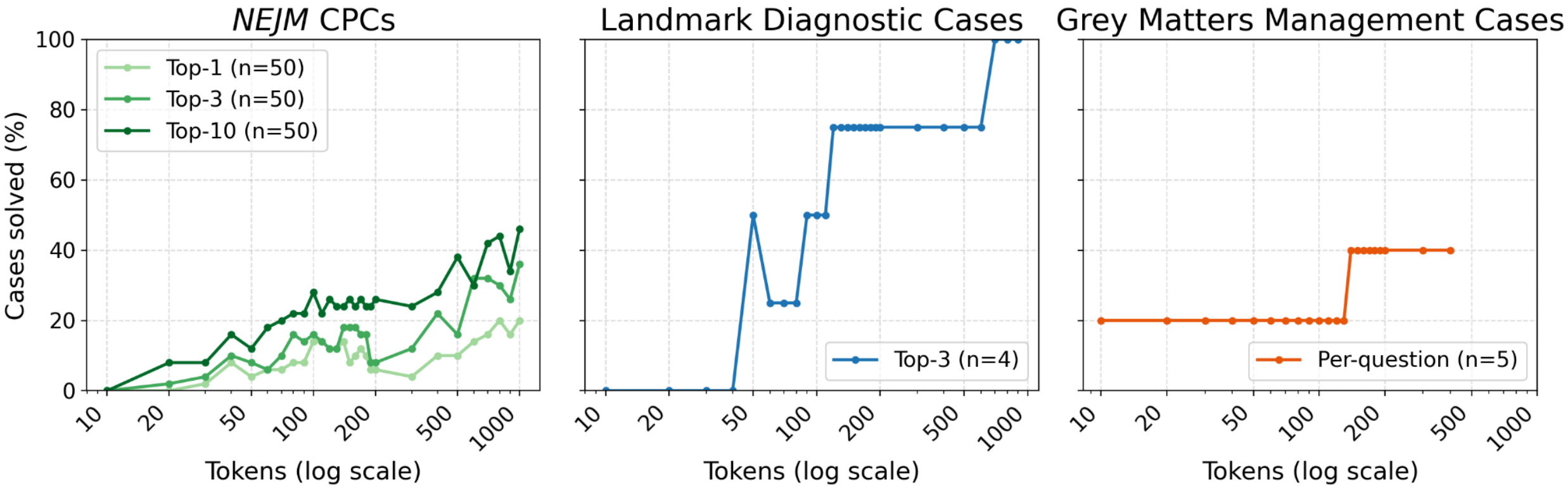

**Fig. 5:** Case-solving performance as a function of how much case text the model sees. Each model was shown progressively longer prefixes of every case, in steps of 10 tokens up to 190 tokens and then steps of 100 tokens up to 1,000 tokens or the end of the case. PrecepTron-32B scored the model's response at each prefix using the task's original rubric. Left: *NEJM* CPCs, showing the fraction of cases in which the true diagnosis appears in the model's top-1, top-3, or top-10 differential (Bond score of 5). Middle: Landmark Diagnostic Cases, showing the fraction of cases scoring at least 80% of the rubric maximum. Right: Grey Matters management questions, showing the fraction scoring at least 80% of the rubric maximum, restricted to the five held-out questions that do not depend on case text revealed in later questions. The x-axis is the number of case-text tokens shown (log scale).

## Box 1: Protocol for deploying an LLM-as-a-judge on a clinical task

**Step 1: Fix the rubric and collect a small physician-scored set.**
Using your new rubric, have physicians score a modest set of responses. On the order of 50–100 scored responses drawn from a handful of cases is a reasonable starting point (e.g., for diagnostic reasoning, our largest gains over the base model came from training on just 93 physician-scored responses spanning 2 cases). Critically, the scored responses should span the quality range: include outputs from weaker and stronger models (and human-written responses where relevant) so that low, middle, and high scores are all represented; a calibration set drawn only from a strong model will contain mostly high scores and teach the judge little about recognizing poor answers. Two or more physicians should rate each response to compute physician inter-rater agreement.

**Step 2: Choose a candidate judge based on your constraints**
If you are working with protected patient data, or hospital policy restricts external APIs, use an open model that runs locally (PrecepTron-32B requires a single 80 GB GPU). If you have no data restrictions and limited computational expertise, a frontier API model is a reasonable starting candidate, but its calibration must still be verified on your task (Step 3). Validation on someone else's task does not transfer. Whenever feasible, evaluate several candidate judges rather than committing to one.

**Step 3: Validate on a test set and screen for known biases.**
Split your cases into a train/test split by case. We used 20% for training and 80% for testing. On your test set check: (1) Agreement: accuracy against the physician score and Cohen's κ. Compare both to physician inter-rater agreement. (2) Confusion matrix: inspect it and check for alignment (3) Inspect a subset of outputs directly (4) Systematic biases: using GRAND-ROUNDS or your own labels, test whether the judge favors longer responses (verbosity bias), AI- over human-written responses, or outputs from its own model family.

**Step 4: If agreement falls short, fine-tune, instead of prompting**
In our comparisons, few-shot prompting and prompt optimization (GEPA) did not consistently improve model performance. Lightweight LoRA fine-tuning on the physician-scored examples was the only strategy that consistently reached physician-level agreement. We release the code on GitHub to achieve this.

**Step 5: Report the judge as part of your methods.**
State the judge model and version, prompts, calibration data, and judge–physician agreement alongside physician inter-rater agreement, and release scores where possible.

**Table 1: Overall Performance of All Models on GRAND-ROUNDS**

| | NEJM CPCs Diagnosis (n=669) | Landmark Diagnostic Cases (n=185) | Grey Matters Management Cases (n=258) | NEJM Healer (n=248) | BIDMC ER (n=719) |
|---|---|---|---|---|---|
| *AI Models* | | | | | |
| Gemma-3-12B | 87% (κ=0.61) | 55% (κ=0.66) | 17% (κ=0.60) | 69% (κ=0.42) | 89% (κ=0.59) |
| Llama-3.1-8B | 75% (κ=0.32) | 43% (κ=0.38) | 44% (κ=0.62) | 46% (κ=0.12) | 75% (κ=0.45) |
| Mistral-3-8B | 53% (κ=0.38) | 54% (κ=0.72) | 62% (κ=0.72) | 63% (κ=0.54) | 75% (κ=0.53) |
| Qwen3.5-9B | 87% (κ=0.62) | 49% (κ=0.43) | 74% (κ=0.83) | 65% (κ=0.47) | 90% (κ=0.65) |
| Qwen3-32B | 82% (κ=0.55) | 46% (κ=0.62) | 72% (κ=0.75) | 76% (κ=0.54) | 87% (κ=0.68) |
| Claude-Opus-4.6 | 74% (κ=0.56) | 28% (κ=0.48) | 84% (κ=0.88) | 82% (κ=0.81) | 91% (κ=0.77) |
| Gemini-3.1-Pro | 83% (κ=0.62) | 68% (κ=0.80) | 80% (κ=0.88) | 78% (κ=0.71) | 86% (κ=0.73) |
| GPT-5 | 87% (κ=0.65) | 30% (κ=0.54) | 71% (κ=0.86) | 75% (κ=0.58) | 87% (κ=0.72) |
| PrecepTron-32B | 92% (κ=0.71) | 61% (κ=0.80) | 80% (κ=0.78) | 81% (κ=0.80) | 91% (κ=0.60) |
| *Physician Baseline (subset of cases scored by at least 2 physicians)* | | | | | |
| Physician | 92% (κ=0.68, n=481) | 67% (κ=0.92, n=115) | 95% (κ=0.90, n=216) | 77% (κ=0.83, n=241) | 88% (κ=0.66, n=719) |

**Table 1:** Performance of several proprietary and open models on the GRAND-ROUNDS benchmark on the held-out test set. Each cell shows accuracy (within-1 on the raw 0–5 Bond scale for NEJM CPCs and BIDMC ER, within-10% of the normalized score elsewhere) and quadratic-weighted Cohen's κ against physician scores. n is the number of test-set entries per

task. For Grey Matters, n is the number of scored cases (258), the unit used for accuracy, corresponding to 1,715 question-level entries.

## Table 2: PrecepTron surfaces the ability of frontier LLMs to generate accurate solutions from 50 or fewer tokens of case text

| Task | Tokens | Text shown to the model | GPT-5 response | Ground truth |
|---|---|---|---|---|
| CPC | 10 | A 19-year-old woman was admitted to this | 1. Anti-NMDA receptor encephalitis (ovarian teratoma–associated)<br>2. Herpes simplex virus type 1 encephalitis<br>3. Autoimmune GABA-A receptor encephalitis | Malignant mixed germ-cell tumor and anti–NMDA receptor encephalitis. |
| CPC | 20 | A 70-year-old woman was evaluated at this hospital because of a racing heart, fatigue, dys | 1. Thyrotoxicosis due to toxic multinodular goiter with atrial fibrillation with rapid ventricular response<br>2. Primary atrial fibrillation with rapid ventricular response (age-related atrial myopathy)<br>3. Atrial flutter with 2:1 AV conduction | Esophageal–pericardial fistula after percutaneous catheter ablation for atrial fibrillation. |
| CPC | 20 | A 63-year-old man was admitted to this hospital with fever, cough, and vision loss in | 1. Giant-cell arteritis (temporal arteritis) with arteritic anterior ischemic optic neuropathy<br>2. Granulomatosis with polyangiitis (GPA) with pulmonary and ocular involvement<br>3. Endogenous bacterial endophthalmitis from pneumonia-associated bacteremia (e.g., Staphylococcus aureus or Klebsiella pneumoniae) | Disseminated infection with hypervirulent Klebsiella pneumoniae. |
| CPC | 30 | A 53-year-old man was evaluated in the cardiology clinic of this hospital because of severe left ventricular dysfunction with an apical aneurys | 1. Ischemic cardiomyopathy from prior anterior MI with true LV apical aneurysm<br>2. Chronic Chagas cardiomyopathy (Trypanosoma cruzi) with LV apical aneurysm<br>3. Hypertrophic cardiomyopathy with midventricular obstruction and apical aneurysm | Chagas' disease. |
| Diagnostic | 20 | History of Present Illness A 58F presents with intermittent episodes of intense itching which have been | 1. Polycythemia vera with aquagenic pruritus<br>2. Primary biliary cholangitis (cholestatic pruritus)<br>3. Hodgkin lymphoma–associated pruritus | Polycythemia Vera |
| Diagnostic | 30 | History of Present Illness A 63F presents to her PCP complaining of progressively worsening fatigue. She says "I just don't feel normal | 1. Primary hypothyroidism<br>2. Iron deficiency anemia due to occult gastrointestinal blood loss<br>3. Major depressive disorder | Pernicious Anemia |
| Diagnostic | 50 | History of Present Illness A 61F is admitted for fever and pain in her eye. Ten days before admission she had a fever. Six days after the fever she had sudden pain in her right hand. It resolved spontaneously. Four days later | 1. Reactive arthritis with acute anterior uveitis<br>2. Disseminated gonococcal infection with ocular involvement<br>3. Giant cell arteritis with ocular ischemia | Infective Endocarditis |
| Diagnostic | 50 | History of Present Illness A 65M comes to his PCP describing a “lack of stamina.” This started eight months ago after his wife died. In addition, he notes a 10-pound weight loss over three months. He has | 1. Major depressive disorder (bereavement-triggered)<br>2. Occult malignancy, particularly colorectal cancer<br>3. Heart failure (new or worsening) | Multiple Myeloma |
| Management | 10 | A 50 year-old woman with alcoholic cirrh<br>Question: In this patient, would you perform a paracentesis and a thoracentesis? What factors push you in one way or another? | Short answer: • Paracentesis: Yes—unless there is a very unusual contraindication. In any hospitalized patient with cirrhosis and ascites, an immediate diagnostic paracentesis is indicated to rule out spontaneous bacteri… | — |
| Management | 10 | A 50 year-old woman with alcoholic cirrh<br>Question: Would you provide any blood products to this patient, either before or after a procedure? Why or why not? | Short answer: - I would not give prophylactic FFP or platelets before a low-bleeding-risk bedside procedure (e.g., diagnostic/therapeutic paracentesis, thoracentesis, central venous line) in a cirrhotic patient, even if… | — |
| Management | 10 | A 55 year-old year-old woman with known<br>Question: How would you evaluate for involuntary weight loss? | Approach overview Unintentional loss of ≥5% of usual body weight over 6–12 months warrants a structured evaluation to identify common, treatable causes (malignancy, endocrine disease, GI disorders including malabsorption… | — |
| Management | 20 | A 72 year-old woman with a history of GERD, type 2 diabetes, and poorly<br>Question: In addition to continuing telemetry monitoring, would you recommend any additional monitoring, testing, or treatment at this time? If so, what? | Yes. With the limited information, the safest next step is to broaden monitoring and obtain targeted tests that reduce immediate risk (arrhythmia/ischemia) and clarify the cause. My reasoning: in a 72-year-old with diabe… | — |

**Table 2:** Representative cases in which GPT-5 reached a high score while shown only a short prefix of the case presentation, scored token-by-token by PrecepTron-32B. For each of three tasks we show the four examples solved with the fewest input tokens (ties broken by case identifier): NEJM clinicopathological conference cases (CPC), Landmark Diagnostic Cases, and Grey Matters Management Cases. A case counts as solved when GPT-5's top-3 differential receives a Bond score of 5 from PrecepTron-32B (CPC) or when the response reaches ≥80% of the rubric maximum (diagnostic and management). Because the diagnostic and management rubrics reward structured clinical reasoning rather than a single correct answer, a high score need not coincide with the listed ground-truth diagnosis;

for diagnostic cases we therefore show GPT-5's three-item differential rather than its final diagnosis. Cases used to train PrecepTron-32B are excluded. Management examples are scored per question, so a single case may contribute more than one row.

## Supplementary Fig. 1: PrecepTron matches physician-level agreement across tasks (Cohen's Kappa)

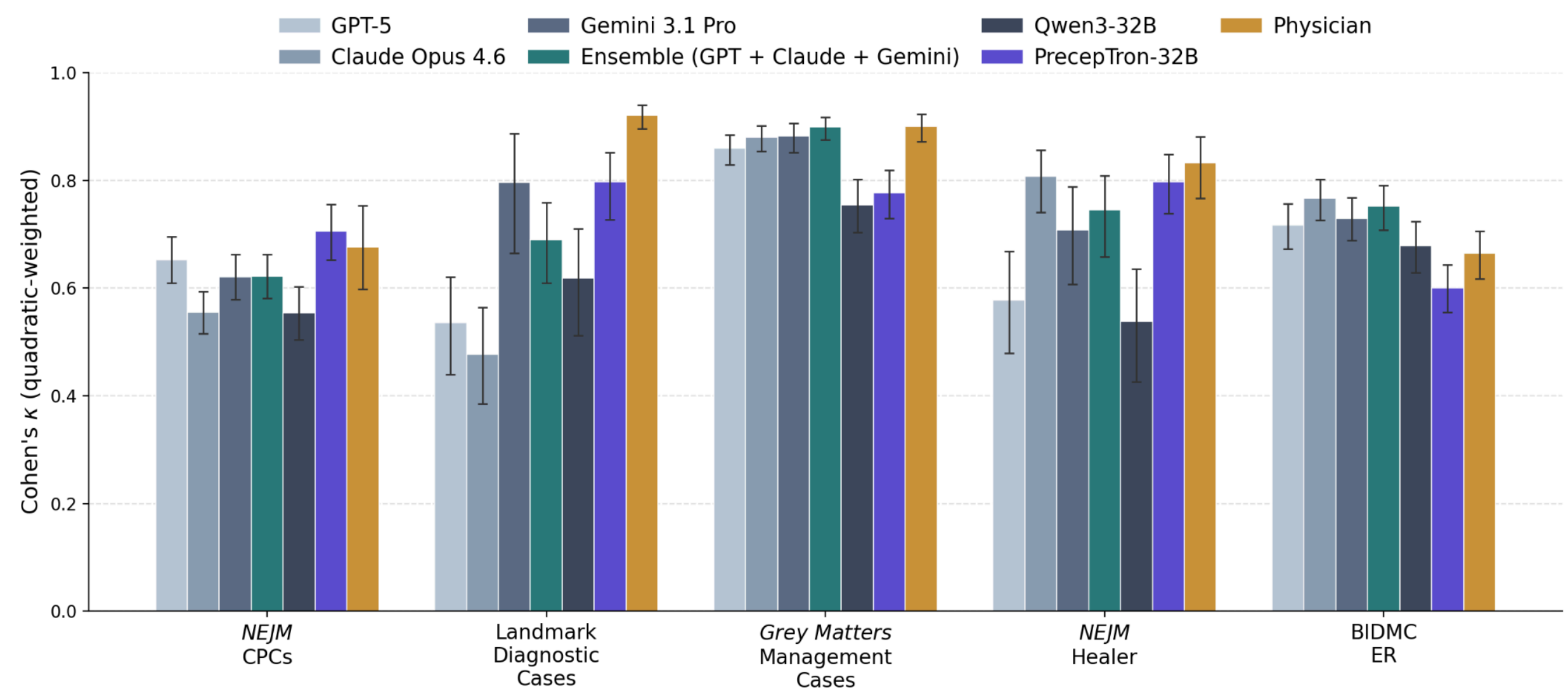

**Supplementary Fig. 1:** Quadratic-weighted Cohen's κ between each judge and the physician score across the five GRAND-ROUNDS tasks. Bars: the three frontier judges (GPT-5, Claude Opus 4.6, Gemini 3.1 Pro), an ensemble of those three, the Qwen3-32B base model, PrecepTron-32B, and the physician inter-rater κ (gold) as the reference. Error bars are 95% bootstrap confidence intervals computed over scored entries.

## Supplementary Fig. 2: LoRA fine-tuning outperforms prompt-based methods

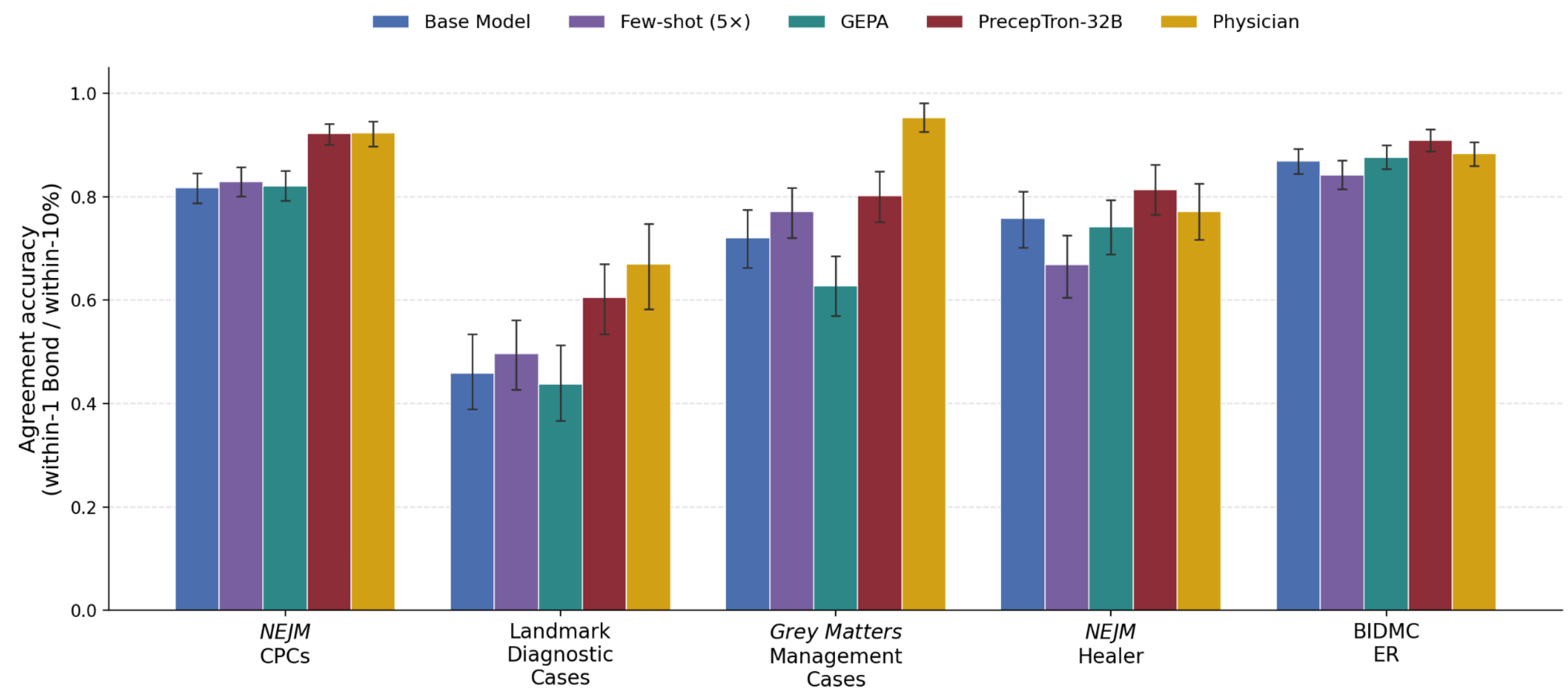

**Supplementary Fig. 2:** Agreement accuracy for the Qwen3-32B judge under four adaptation strategies: base model, few-shot prompting (five physician-scored examples), GEPA prompt optimization, and PrecepTron-32B (LoRA fine-tuning). Error bars are 95% bootstrap confidence intervals computed over scored entries.

## Supplementary Fig. 3: Retraining PrecepTron without AI examples in Landmark Diagnostic Cases

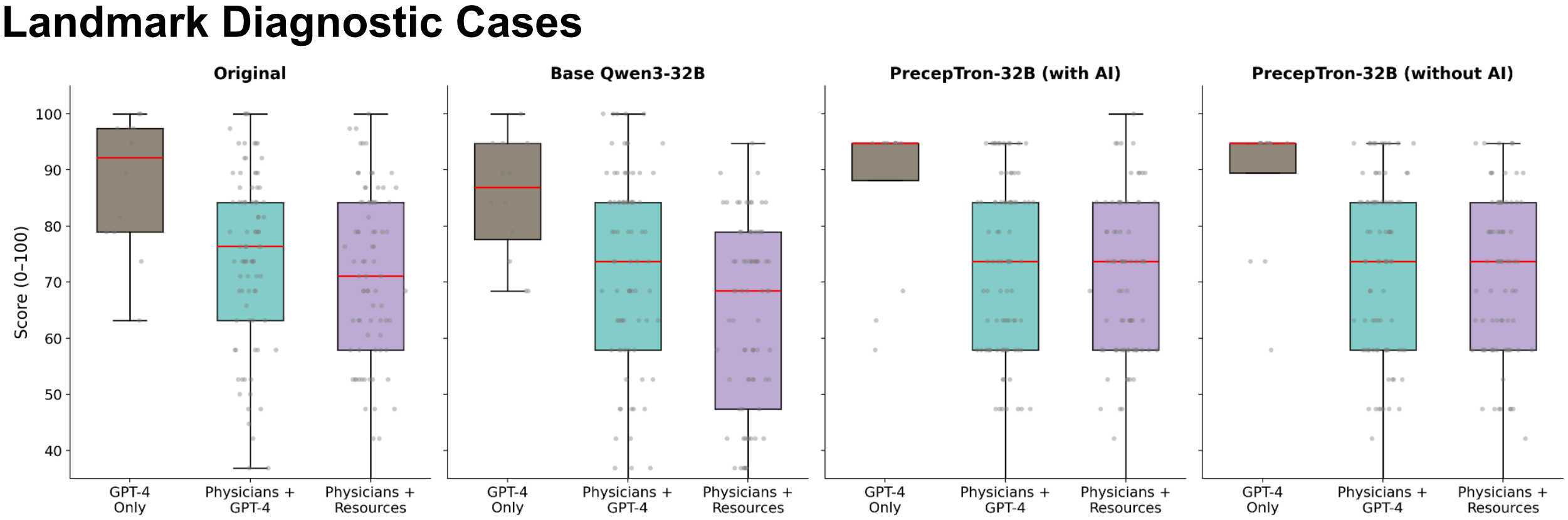

**Supplementary Fig. 3:** Held-out Landmark diagnostic-reasoning scores (0–100% of the 19-point rubric) by response source for four graders: the original physician scores, the Base Qwen3-32B judge, PrecepTron-32B trained with the AI-authored (o1, GPT-4) examples, and the same recipe and seed trained without them. Boxes show median and IQR; points are individual responses.

## Supplementary Fig. 4: PrecepTron-32B agreement with physicians

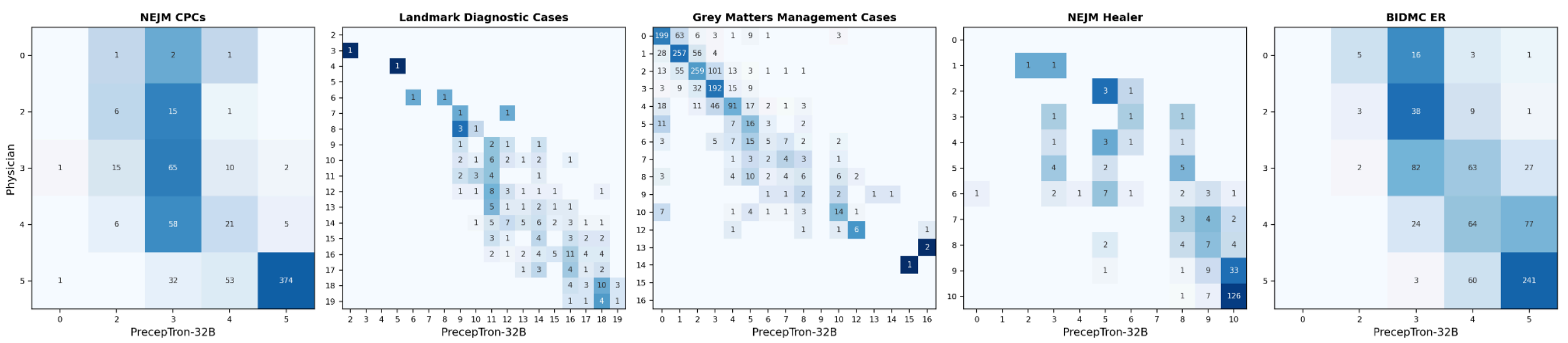

**Supplementary Fig. 4:** Per-task confusion matrices of PrecepTron-32B (primary seed) against the physician score on held-out cases. Rows are the physician score and columns are the PrecepTron score.

## Supplementary Fig. 5: LLM-judge Model Preferences Across Four Clinical-Reasoning Tasks

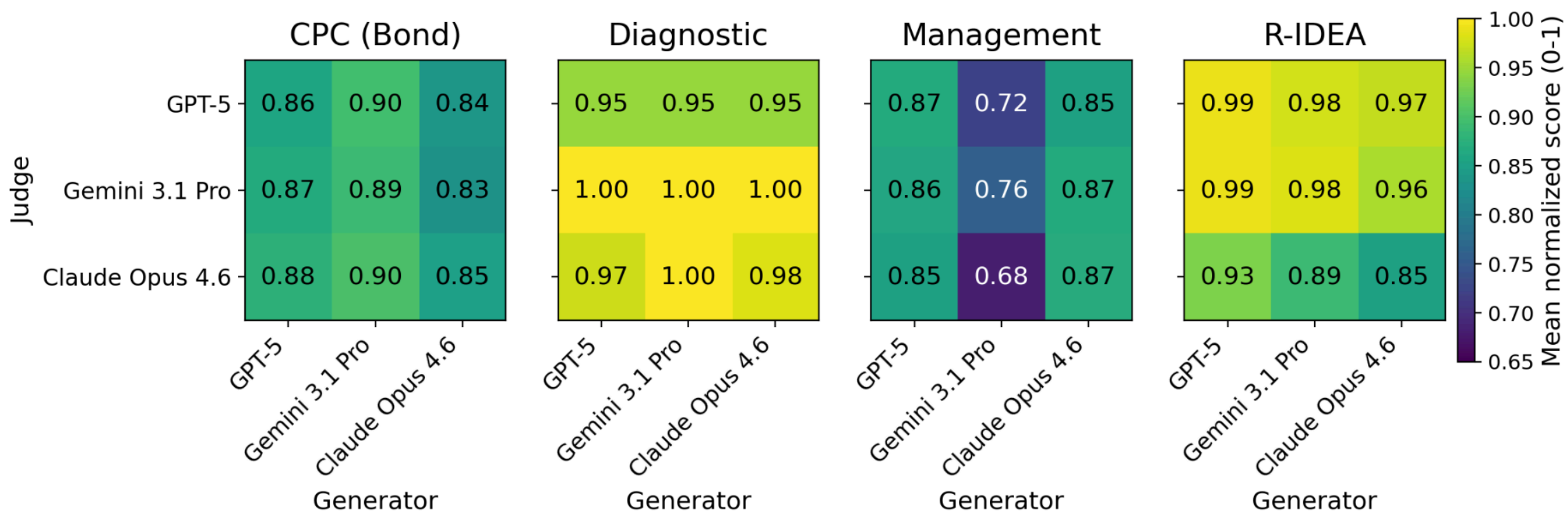

**Supplementary Fig. 5:** Mean LLM-judge scores by response generator and judge across four clinical-reasoning tasks (CPC/Bond, Diagnostic, Management, R-IDEA). Three frontier models, GPT-5, Gemini 3.1 Pro, and Claude Opus 4.6, each generated responses to every case using the original study prompts. For the *NEJM* CPCs, we prompted models to provide at most 3 diagnoses. Each model then scored all responses as an LLM judge using the corresponding task rubric. Rows are judges, columns are generators; the red outline marks self-judging cells (judge and generator are the same model). Cell values are the mean score normalized to each task's maximum (CPC /5, Diagnostic /19, Management /per-question rubric max, R-IDEA /10), averaged over all cases in the task (CPC n=143, Diagnostic n=6, Management n=33 standardized questions across all five cases, R-IDEA n=80 case-aliquots).

## Supplementary Table 1: LoRA fine-tuning of Qwen3-32B with different number of epochs and resampling

| | NEJM CPCs Diagnosis | Landmark Diagnostic Cases | Grey Matters Management Cases | NEJM Healer | BIDMC ER |
|---|---|---|---|---|---|
| SFT 1ep (unbalanced) | 92% ±2% (κ=0.72 ±0.03) | 57% ±4% (κ=0.77 ±0.01) | 71% ±2% (κ=0.76 ±0.00) | 75% ±2% (κ=0.57 ±0.19) | 89% ±1% (κ=0.62 ±0.03) |
| SFT 3ep (unbalanced) | 95% ±1% (κ=0.74 ±0.01) | 64% ±4% (κ=0.81 ±0.01) | 77% ±2% (κ=0.78 ±0.01) | 77% ±2% (κ=0.51 ±0.06) | 93% ±1% (κ=0.63 ±0.03) |
| SFT 1ep (balanced) | 89% ±1% (κ=0.71 ±0.02) | 54% ±3% (κ=0.79 ±0.02) | 77% ±3% (κ=0.79 ±0.00) | 76% ±3% (κ=0.69 ±0.03) | 89% ±2% (κ=0.64 ±0.01) |
| PrecepTron-32B | 93% ±2% (κ=0.73 ±0.03) | 60% ±0% (κ=0.81 ±0.01) | 79% ±2% (κ=0.76 ±0.02) | 79% ±2% (κ=0.72 ±0.06) | 91% ±1% (κ=0.63 ±0.03) |

**Supplementary Table 1:** Agreement accuracy and quadratic-weighted Cohen's κ against the physician score (each mean ± standard deviation across three LoRA seeds) for the 1-epoch vs 3-epoch and balanced vs unbalanced resampling configurations. PrecepTron-32B is the balanced 3-epoch configuration.